%% file: main.tex
\documentclass[10pt,twocolumn,letterpaper]{article}

\usepackage[pagenumbers]{cvpr} 
\usepackage{indentfirst}     
\usepackage{algorithm}
\usepackage{algpseudocode}
\usepackage{xcolor}
\usepackage{colortbl}
\usepackage{multirow}
\usepackage{graphicx}
\usepackage{amsmath}
\usepackage{amsthm}
\usepackage{caption}

\definecolor{cvprblue}{rgb}{0.21,0.49,0.74}
\definecolor{mydarkgreen}{RGB}{0, 128, 0}
\usepackage[pagebackref,breaklinks,colorlinks,citecolor=cvprblue]{hyperref}

\def\paperID{3236} 
\def\confName{CVPR}
\def\confYear{2026}

\title{HookMIL: Revisiting Context Modeling in Multiple Instance Learning for Computational Pathology}

\author{Xitong Ling\textsuperscript{1*}, Minxi Ouyang\textsuperscript{1*}, Xiaoxiao Li\textsuperscript{1*}, Jiawen Li\textsuperscript{1},
Ying Chen\textsuperscript{2}, Yuxuan Sun\textsuperscript{3}, \\ Xinrui Chen\textsuperscript{1},   Tian Guan\textsuperscript{1} Xiaoping Liu\textsuperscript{4\dag}, Yonghong He\textsuperscript{1\dag}\\
\textsuperscript{1}Shenzhen International Graduate School, Tsinghua University\\
\textsuperscript{2}School of Informatics, Xiamen University\\
\textsuperscript{3}School of Engineering,  Westlake University\\
\textsuperscript{4}Zhongnan Hospital,  Wuhan University\\
{\tt\small \{lingxt23, oymx23, lixiaoxiao25\}@mails.tsinghua.edu.cn}\\{\tt\small liuxiaoping@whu.edu.cn}, {\tt\small heyh@sz.tsinghua.edu.cn} \\
}

\begin{document}
\maketitle
\input{0_abstract}

{\let\thefootnote\relax\footnote{* Contributed equally. \dag Corresponding authors.}}
\input{1_intro}

\input{2_relatedwork}
\input{3_methodology}
\input{4_experiments}
\input{5_conclusion}
{
    \small
    \bibliographystyle{ieeenat_fullname}
    \bibliography{main}
}

\input{6_supp}

\end{document}

%% file: 0_abstract.tex
\begin{abstract}
Multiple Instance Learning (MIL) has enabled weakly supervised analysis of whole-slide images (WSIs) in computational pathology. However, traditional MIL approaches often lose crucial contextual information, while transformer-based variants, though more expressive, suffer from quadratic complexity and redundant computations. To address these limitations, we propose \textbf{HookMIL}, a context-aware and computationally efficient MIL framework that leverages compact, learnable hook tokens for structured contextual aggregation. These tokens can be initialized from (i) key-patch visual features, (ii) text embeddings from vision--language pathology models, and (iii) spatially grounded features from spatial transcriptomics--vision models. This multimodal initialization enables Hook Tokens to incorporate rich textual and spatial priors, accelerating convergence and enhancing representation quality. During training, Hook tokens interact with instances through bidirectional attention with linear complexity. To further promote specialization, we introduce a Hook Diversity Loss that encourages each token to focus on distinct histopathological patterns. Additionally, a hook-to-hook communication mechanism refines contextual interactions while minimizing redundancy. Extensive experiments on four public pathology datasets demonstrate that HookMIL achieves state-of-the-art performance, with improved computational efficiency and interpretability.
Codes are available at \url{https://github.com/lingxitong/HookMIL}.

\end{abstract}

%% file: 1_intro.tex
\section{Introduction}
\label{sec:intro}
\begin{figure}[t]
\centering
\includegraphics[width=1\columnwidth]{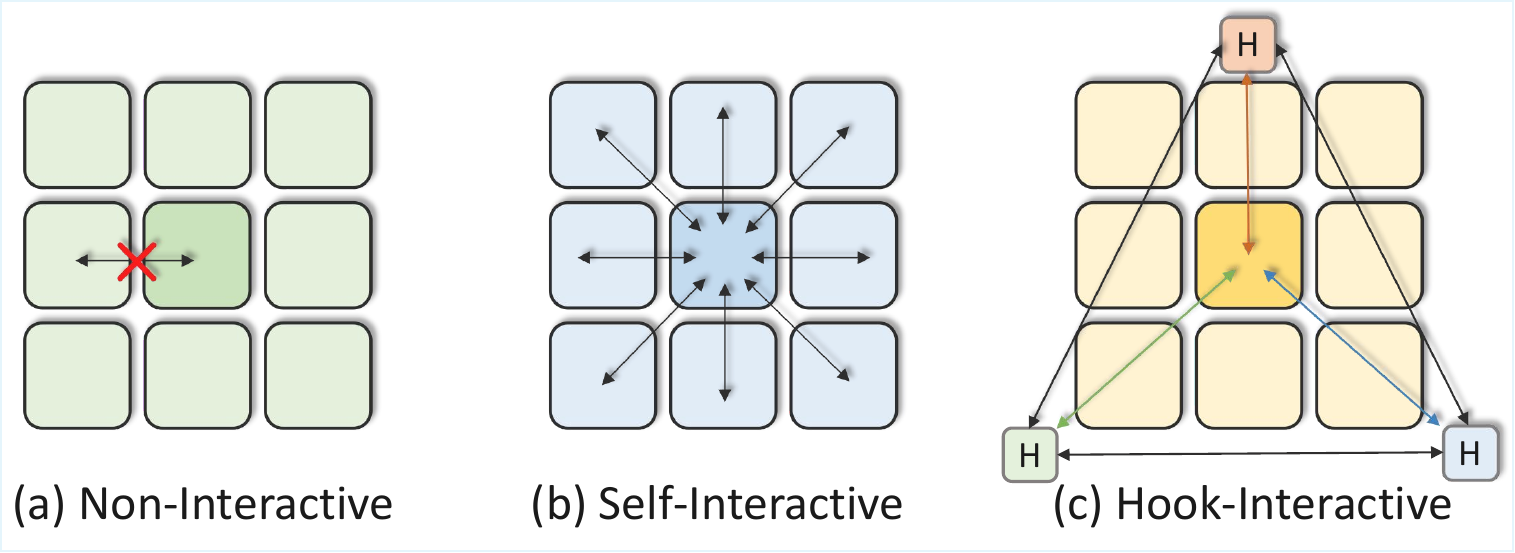}
\caption{Comparison of different context modeling paradigms in MIL: (a) non-interactive context modeling, (b) self-interactive context modeling, and (c) hook-interactive context modeling.}
\label{fig1}
\end{figure}

Whole-slide images (WSIs) have transformed histopathological diagnosis by enabling the high-resolution digitization of entire tissue sections at micrometer resolution. A single WSI can encompass over \(100{,}000 \times 100{,}000\) pixels, capturing a wealth of multiscale information ranging from nuclear morphology to tissue architecture and spatial organization patterns. This hierarchical information is essential for accurate cancer diagnosis, grading, and prognosis prediction, as pathologists routinely evaluate not only individual cellular features but also their spatial relationships, tissue context, and architectural patterns, all of which are crucial for defining disease states.

The computational analysis of WSIs poses unique challenges due to their vast size and the limited availability of detailed annotations. Manual pixel-level or region-level annotation is not only prohibitively expensive but also time-consuming, requiring highly specialized expertise. To overcome these challenges, Multiple Instance Learning (MIL) has emerged as the dominant framework, treating each WSI as a ``bag'' of instances (image patches) with only slide-level labels available during training \cite{chen2022scaling, wang2018revisiting}. This weakly supervised paradigm has driven significant progress in computational pathology, enabling advances in cancer detection, subtyping, and outcome prediction.

Despite their widespread adoption, conventional MIL approaches suffer from a fundamental limitation: they typically process each instance independently through instance-wise feature extraction and aggregation, ignoring the rich contextual relationships between spatially adjacent or morphologically similar regions. This loss of context is particularly problematic in computational pathology, where diagnostic decisions often depend critically on understanding tissue architecture, spatial patterns, and neighborhood relationships. For example, identifying tumor invasion requires analyzing the complex interface between malignant and benign regions, while grading tumor aggressiveness involves assessing cellular organization patterns and stromal interactions. The diagnosis of certain cancer subtypes may depend on the recognition of specific spatial arrangements of cells that cannot be discerned from isolated instances.

Recent Transformer-based MIL methods \cite{shao2021transmil, chikontwe2024fr} attempt to address this limitation by employing global self-attention mechanisms to model interactions between all instance pairs. However, native Transformer-based modeling introduces substantial information redundancy, and approximate Transformer-based instance context modeling suffers from the same issue. Although these approaches improve performance by capturing long-range dependencies, they introduce a significant drawback: indiscriminate modeling of all pairwise interactions, including many irrelevant or redundant connections that may dilute important diagnostic signals and introduce noise.

Moreover, existing attention mechanisms in MIL often lack interpretable specialization, with different attention heads potentially learning overlapping or redundant patterns. This reduces the transparency of the model and makes it difficult to correlate learned representations with known pathological concepts. As computational pathology moves toward clinical deployment, there is a growing need for methods that are not only accurate but also efficient, scalable, and interpretable. 

To overcome these limitations, we propose \textbf{HookMIL}, a novel MIL framework that introduces a lightweight yet powerful context modeling mechanism through learnable hook tokens. Our key insight is that meaningful contextual dependencies in WSIs can be effectively captured through a small set of latent anchors that selectively aggregate and redistribute information, rather than exhaustively connecting all instance pairs. This approach is inspired by how pathologists interpret tissue samples: they do not examine every cell--cell relationship individually but instead identify key morphological patterns and architectural features that serve as diagnostic anchors.

HookMIL employs a bidirectional attention mechanism, where hooks first aggregate information from relevant instances through cross-attention, then communicate with each other to model higher-order dependencies, and finally project enhanced contextual signals back to the instance space. This creates an efficient information flow that preserves important contextual relationships while avoiding the computational burden of global self-attention. Additionally, we introduce a novel hook diversity loss, which explicitly encourages different hooks to focus on distinct morphological patterns, thus enhancing both model interpretability and representational power.

The development of multimodal pathology pretraining models further provides a flexible hook initialization mechanism for HookMIL, allowing it to effectively integrate diverse data modalities and adapt to the complexity of real-world diagnostic tasks.

The main contributions of this work are:
\begin{itemize}
\item We propose HookMIL, a novel hook-based context-aware MIL framework that models instance dependencies through learnable hook tokens, enabling efficient and scalable context modeling for large WSIs.
\item We introduce a Hook Diversity Loss that encourages hooks to specialize in distinct morphological patterns, improving model interpretability and preventing representation collapse.
\item We provide theoretical analysis demonstrating HookMIL's gradient propagation properties and its interpretation as learning a low-rank basis for inter-instance dependencies.
\item Extensive experiments on four public pathology datasets show that HookMIL achieves state-of-the-art performance while significantly reducing computational overhead compared to Transformer-based alternatives.
\end{itemize}

%% file: 2_relatedwork.tex
\section{Related Work}
\label{sec:related}

\subsection{Multiple Instance Learning}
MIL has emerged as the standard paradigm for weakly supervised WSI analysis, where a slide (bag) is represented by a set of patch embeddings (instances) and only slide-level labels are available. Early attention-based MIL methods, such as ABMIL \cite{ilse2018attention}, learn an instance-weighting function to aggregate patch features into a slide descriptor without explicitly modeling inter-instance relationships. Subsequent variants introduce architectural refinements (e.g., gated or multi-head attention), curriculum strategies, and robust pooling schemes, but generally maintain the assumption that instances are encoded independently and only mixed at the aggregation stage. Dual-stream designs (e.g., DSMIL \cite{li2021dual}) further couple an instance classifier with an attention head to stabilize optimization and highlight critical instances, yet still avoid explicitly modeling pairwise dependencies among all patches. Cluster-aware attention MIL (e.g., CLAM \cite{lu2021data}) enhances interpretability by discovering sub-bag clusters while maintaining a largely instance-independent aggregation scheme.

\paragraph*{Non-interactive vs.\ Interactive MIL.}
As shown in Fig.~\ref{fig1}, from the perspective of whether instances directly interact before aggregation, MIL methods can be grouped into two families: \emph{non-interactive} and \emph{interactive}. \textbf{Non-interactive MIL} (e.g., ABMIL \cite{ilse2018attention}, DSMIL \cite{li2021dual}, CLAM \cite{lu2021data}) computes attention weights over independently encoded instances and aggregates them with a learned pooling operator; no explicit message passing occurs among instances prior to pooling. This family is attractive for its simplicity and linear scaling in the number of instances, but it may suffer from context loss because morphological patterns are often defined by local neighborhoods rather than isolated tiles.

In contrast, \textbf{interactive MIL} explicitly establishes dependencies among instances before aggregation. Transformer-based MIL (e.g., TransMIL \cite{shao2021transmil}, FRMIL \cite{chikontwe2024fr}) builds full self-attention over the bag to capture long-range relationships. Hierarchical Transformers (e.g., HIPT-like designs \cite{chen2022scaling}) introduce multi-stage tokenization to reduce the quadratic cost while retaining context by first fusing local neighborhoods and then propagating global information. Graph-based MIL methods \cite{li2024dynamic, chen2021whole} construct slide graphs where nodes are patches and edges encode spatial proximity or feature similarity; graph convolution or graph attention then propagates messages to model tissue topology prior to aggregation. Several works also apply linear/sparse attention (e.g., Performer-/Nyström-style approximations \cite{shao2021transmil, chikontwe2024fr}) or top-$k$ sparsification to scale interactive MIL to large bags by limiting pairwise computations. Although interactive MIL excels at maintaining contextual coherence, it also incurs notable redundancy and often has limited capacity to model diagnostically significant relationships in a focused manner.

\subsection{Computational Pathology Foundation Models}
Recent pathology foundation models (PFMs) \cite{wang2022transformer,chen2024towards,vorontsov2024foundation,xu2024whole,zhu2025subspecialty} have markedly improved the quality and transferability of pathology-domain representations. Self-supervised visual encoders trained on millions of pathology images with contrastive or redundancy-reduction objectives, including MoCo \cite{he2020momentum}, DINO \cite{caron2021emerging}, and DINOv2 \cite{oquab2023dinov2}, deliver robust and stain-tolerant features that generalize across organs and institutions. These encoders substantially boost downstream MIL performance with minimal task-specific supervision. 

Another line of research focuses on leveraging additional modalities paired with pathology images to develop multimodal foundation models based on architectures similar to CLIP \cite{radford2021learning} or CoCa \cite{yu2022coca}. Representative approaches include models pretrained on image--text pairs, such as PLIP \cite{huang2023visual} and CONCH \cite{lu2024visual}, as well as models integrating images with spatial transcriptomics data, such as Loki \cite{chen2025visual}. These multimodal foundation models align visual representations with other modality-specific embeddings, thereby not only providing powerful visual encoders for downstream tasks but also enabling the incorporation of prior knowledge through cross-modal alignment. This offers a promising pathway for introducing external semantic information into pathology-related applications.

%% file: 3_methodology.tex
\section{Methodology}
\label{sec:method}

\begin{figure*}[t]
\centering
\includegraphics[width=\textwidth]{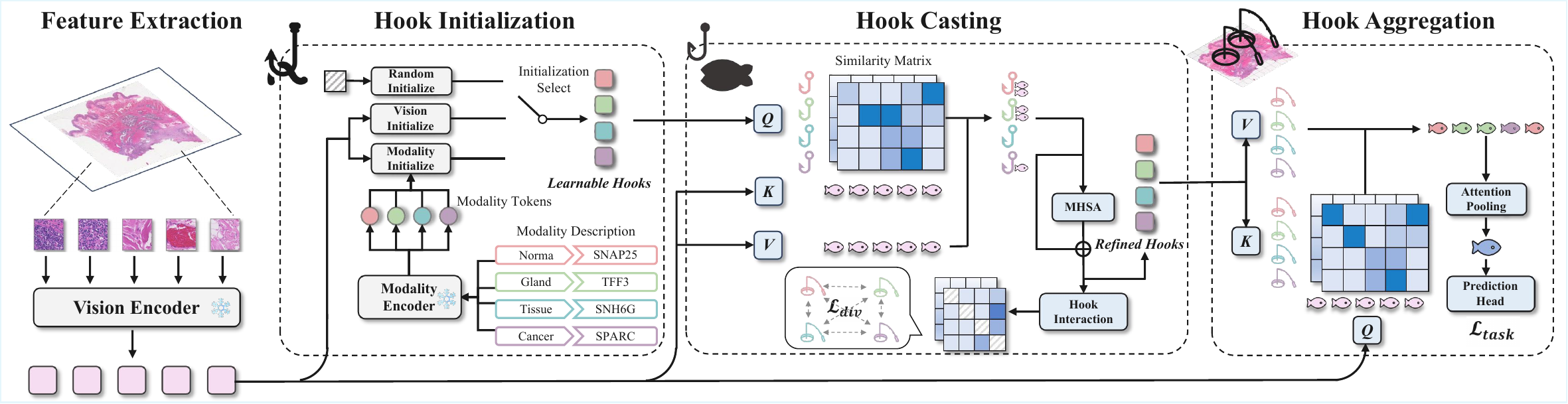}
\caption{Overall framework of the proposed HookMIL for WSI analysis, including patch feature extraction, hook initialization, hook casting, and hook aggregation.}
\label{fig2}
\end{figure*}

\subsection{Problem Formulation}

In Multiple Instance Learning (MIL) for computational pathology, each whole-slide image (WSI) is treated as a bag containing numerous instances (patches). Let $X = \{x_1, x_2, \ldots, x_N\}$ represent a bag with $N$ instances, where each instance $x_i \in \mathbb{R}^{D}$ corresponds to a patch-level feature vector. The bag-level label $y \in \{0, 1, \ldots, C-1\}$ indicates the overall diagnostic category. The key challenge is to learn a mapping $f(X) \rightarrow y$ that accurately predicts the bag-level label while only having access to weak supervision at the bag level.

Traditional attention-based MIL methods compute instance importance weights using a simple attention mechanism:
\begin{equation}
    a_i = \frac{\exp(w^{\top} \tanh(V x_i))}{\sum_{j=1}^N \exp(w^{\top} \tanh(V x_j))},
    \label{eq:traditional_attention}
\end{equation}
\begin{equation}
    z = \sum_{i=1}^N a_i x_i,
    \label{eq:traditional_aggregation}
\end{equation}
where $w \in \mathbb{R}^{D_a}$ and $V \in \mathbb{R}^{D_a \times D}$ are learnable parameters, $D_a$ is the attention dimension, and $z \in \mathbb{R}^{D}$ is the aggregated bag-level representation. However, this approach processes each instance independently, ignoring the rich contextual relationships between morphologically similar regions across the slide.

\subsection{HookMIL Architecture}

HookMIL introduces a novel contextual modeling framework that captures inter-instance dependencies through learnable hook tokens. The overall architecture operates in four sequential phases as illustrated in Figure~\ref{fig2}. For a more concrete description of the overall workflow of HookMIL, please refer to the pseudo-code in Algorithm~\ref{alg:algorithm1} in the Supplementary Material.

\subsubsection{Hook Token Initialization}

We introduce $K$ learnable hook tokens $H = [h_1, h_2, \ldots, h_K]^\top \in \mathbb{R}^{K \times D}$, where $K \ll N$. These tokens are initialized using a truncated normal distribution with a standard deviation of 0.02:
\begin{equation}
    h_k \sim \mathcal{TN}(0, 0.02^2) \quad \text{for } k = 1, 2, \ldots, K,
    \label{eq:hook_initialization}
\end{equation}
where $\mathcal{TN}$ denotes the truncated normal distribution. The hook tokens serve as adaptive contextual anchors that learn to represent prototypical morphological patterns.

In practice, this initialization scheme is flexible and can be adapted to different types of foundation models, giving rise to four variants of our approach, denoted as \textbf{HookMIL-X}, where $X$ specifies the initialization strategy:
    


\begin{itemize}
    \item \textbf{HookMIL-TN}: Default variant, where hook tokens are randomly initialized from the truncated normal distribution in Eq.~\ref{eq:hook_initialization}. This model-agnostic setting serves as a baseline.
    
    \item \textbf{HookMIL-Vis}: For visual pathology foundation models, hook tokens are initialized from representative visual embeddings of prototypical lesion regions in the training set, injecting visual inductive bias and potentially accelerating convergence.

    \item \textbf{HookMIL-Txt}: With vision--language models (e.g., pretrained image--text encoders), hook tokens are initialized from textual embeddings of pathology-specific knowledge (e.g., disease terms, tissue types, diagnostic phrases), introducing semantic priors via language grounding.

    \item \textbf{HookMIL-ST}: In multimodal settings with spatial transcriptomics (e.g., image--ST foundation models), hook tokens are initialized from embeddings of gene expression profiles or molecular descriptors, allowing molecular information to guide visual representation learning.
\end{itemize}

This flexible initialization mechanism allows the hook tokens to encode prior knowledge from various sources, facilitating better generalization, interpretability, and convergence behavior in downstream MIL tasks.

\subsubsection{Hook-to-Instance Aggregation}

In the first attention phase, hook tokens query the instance features to gather relevant contextual information. Given input instances $X = [x_1, x_2, \ldots, x_N]^\top \in \mathbb{R}^{N \times D}$, we compute query, key, and value projections:
\begin{equation}
    Q_h = H W_q^\top, \quad K_x = X W_k^\top, \quad V_x = X W_v^\top,
    \label{eq:hook_to_instance_projections}
\end{equation}
where $W_q, W_k, W_v \in \mathbb{R}^{D \times D}$ are learnable weight matrices. Attention weights are computed as:
\begin{equation}
    A_{h2x} = \mathrm{softmax}\left(\frac{Q_h K_x^\top}{\sqrt{D}}\right) \in \mathbb{R}^{K \times N},
    \label{eq:hook_to_instance_attention}
\end{equation}
where the softmax function is applied row-wise. Context aggregation and token update are performed as:
\begin{equation}
    \tilde{H} = \mathrm{LayerNorm}(H + A_{h2x} V_x) \in \mathbb{R}^{K \times D}.
    \label{eq:hook_update}
\end{equation}
The resulting updated hook tokens $\tilde{H}$ now encode contextual information from semantically related instances.

\subsubsection{Hook Intercommunication}

To model relationships between different contextual patterns, we enable hook tokens to exchange information through multi-head self-attention:
\begin{equation}
    H_{\text{inter}} = \mathrm{MHSA}(\tilde{H}, \tilde{H}, \tilde{H}),
    \label{eq:mhsa}
\end{equation}
\begin{equation}
    H' = \mathrm{LayerNorm}(\tilde{H} + H_{\text{inter}}) \in \mathbb{R}^{K \times D},
    \label{eq:hook_final}
\end{equation}
where $\mathrm{MHSA}$ denotes multi-head self-attention with $L$ heads. This phase allows hooks to refine their representations based on complementary information from other hooks.

\subsubsection{Instance-to-Hook Feedback}

In the final phase, the refined hook tokens propagate contextual information back to the instance representations. We compute:
\begin{equation}
    Q_x = X W_q'^\top, \quad K_h = H' W_k'^\top, \quad V_h = H' W_v'^\top,
    \label{eq:instance_to_hook_projections}
\end{equation}
where $W_q', W_k', W_v' \in \mathbb{R}^{D \times D}$ are learnable weight matrices. Attention weights are computed as:
\begin{equation}
    A_{x2h} = \mathrm{softmax}\left(\frac{Q_x K_h^\top}{\sqrt{D}}\right) \in \mathbb{R}^{N \times K},
    \label{eq:instance_to_hook_attention}
\end{equation}
where the softmax function is applied row-wise. The instance representations are updated as:
\begin{equation}
    X' = \mathrm{LayerNorm}(X + A_{x2h} V_h) \in \mathbb{R}^{N \times D}.
    \label{eq:instance_update}
\end{equation}
The output $X'$ contains context-aware instance representations where each instance is enhanced with relevant morphological context.

\subsection{Diversity Regularization}

To prevent hook tokens from collapsing to similar representations, we introduce a diversity regularization loss. Let $L \in \mathbb{R}^{K \times N}$ be the hook-to-instance attention logits from Equation~\ref{eq:hook_to_instance_attention} before softmax normalization. We first normalize these logits:
\begin{equation}
    L_{\text{norm}} = \frac{L}{\|L\|_F + \epsilon},
    \label{eq:logit_normalization}
\end{equation}
where $\|\cdot\|_F$ denotes the Frobenius norm and $\epsilon = 10^{-6}$ is a small constant for numerical stability. The similarity matrix between hook attention distributions is computed as:
\begin{equation}
    S = L_{\text{norm}} L_{\text{norm}}^\top \in \mathbb{R}^{K \times K}.
    \label{eq:similarity_matrix}
\end{equation}
The diversity loss specifically penalizes similarities between different hooks:
\begin{equation}
    \mathcal{L}_{\text{div}} = \frac{1}{K(K-1)} \sum_{i=1}^K \sum_{j\neq i} S_{ij}^2,
    \label{eq:diversity_loss}
\end{equation}

\subsection{Training Objective}

The complete HookMIL architecture integrates the hook mechanism with standard MIL components. The context-aware instance representations and diversity loss are obtained as:
\begin{equation}
    X', \mathcal{L}_{\text{div}} = \mathrm{HookMIL}(X),
    \label{eq:hookmil_output}
\end{equation}
We then perform attention-based pooling to obtain the bag-level representation:
\begin{equation}
    a_i = \frac{\exp(w_a^\top \tanh(V_a x'_i))}{\sum_{j=1}^N \exp(w_a^\top \tanh(V_a x'_j))},
    \label{eq:enhanced_attention}
\end{equation}
\begin{equation}
    z = \sum_{i=1}^N a_i x'_i \in \mathbb{R}^{D},
    \label{eq:enhanced_aggregation}
\end{equation}
where $w_a \in \mathbb{R}^{D_a}$ and $V_a \in \mathbb{R}^{D_a \times D}$ are learnable parameters. The bag-level prediction is obtained as:
\begin{equation}
    \hat{y} = \mathrm{softmax}(W_c z + b_c) \in \mathbb{R}^{C},
    \label{eq:classification}
\end{equation}
where $W_c \in \mathbb{R}^{C \times D}$ and $b_c \in \mathbb{R}^{C}$ are the classifier parameters. The overall training objective combines the task-specific loss with diversity regularization:
\begin{equation}
    \mathcal{L}_{\text{total}} = \mathcal{L}_{\text{ce}}(\hat{y}, y) + \lambda \mathcal{L}_{\text{div}},
    \label{eq:total_loss}
\end{equation}
where $\mathcal{L}_{\text{ce}}$ is the cross-entropy loss for classification, and $\lambda$ is a weighting coefficient that controls the contribution of the diversity loss.

\subsection{Computational Efficiency}

The hierarchical attention design of HookMIL provides significant computational advantages over global self-attention. Let $N$ be the number of instances, $K$ the number of hooks, and $D$ the feature dimension. The computational complexities of each phase are:

\begin{itemize}
    \item \textbf{Hook-to-Instance Aggregation}: $\mathcal{O}(NKD)$ complexity
    \item \textbf{Hook Intercommunication}: $\mathcal{O}(K^2D)$ complexity  
    \item \textbf{Instance-to-Hook Feedback}: $\mathcal{O}(NKD)$ complexity
\end{itemize}

The overall complexity is $\mathcal{O}(NK + K^2)$, which is linear in $N$ when $K$ is fixed. For typical WSI analysis with $N=10,000$ instances and $K=8$ hooks, HookMIL reduces computational complexity by approximately $1250\times$ compared to global self-attention's $\mathcal{O}(N^2)$ complexity, while effectively capturing long-range dependencies between instances.

\subsection{Theoretical Analysis}
We further analyze the hook mechanism from the perspectives of low-rank structure and gradient behavior; detailed discussions are provided in Section~\ref{supp_low_rank} and Section~\ref{supp_grid} of Supplementary Material.

%% file: 4_experiments.tex
\section{Experiments}
\definecolor{customcolor}{rgb}{0.9176, 0.6667, 1.0000}  

\begin{table*}[htbp]
\caption{Performance [\%] of different MIL methods on multiple WSI diagnostic classification tasks based on UNI \cite{chen2024towards}.}
\label{table1}
\centering
\resizebox{2\columnwidth}{!}{
\renewcommand{\arraystretch}{1.5}{
\begin{tabular}{lccccccccc}
\hline
\multirow{2}{*}{\textbf{Method}} &
  \multicolumn{3}{c}{\textbf{CAMELYON$^{+}$ \cite{ling2025comprehensive}}} &
  \multicolumn{3}{c}{\textbf{PANDA \cite{bulten2022artificial}}} &
  \multicolumn{3}{c}{\textbf{TISSUENET \cite{lomenie2022can}}} \\ \cline{2-10} 
 &
  \textbf{Accuracy} &
  \textbf{AUC} &
  \textbf{F1-score} &
  \textbf{Accuracy} &
  \textbf{AUC} &
  \textbf{F1-score} &
  \textbf{Accuracy} &
  \textbf{AUC} &
  \textbf{F1-score} \\ \hline
ABMIL \cite{ilse2018attention} &
  $93.40_{1.45}$ &
  $96.76_{0.75}$ &
  \multicolumn{1}{c|}{$92.85_{1.48}$} &
  $72.64_{0.66}$ &
  $93.41_{0.18}$ &
  \multicolumn{1}{c|}{$67.92_{1.08}$} &
  $73.84_{1.72}$ &
  $92.37_{0.73}$ &
  $74.42_{1.67}$ \\
CLAM-SB \cite{lu2021data} &
  $93.10_{1.12}$ &
  $96.92_{0.47}$ &
  \multicolumn{1}{c|}{$92.39_{1.28}$} &
  $73.03_{1.26}$ &
  $93.70_{0.26}$ &
  \multicolumn{1}{c|}{$68.53_{1.66}$} &
  $74.14_{1.44}$ &
  $93.04_{0.74}$ &
  $74.60_{1.30}$ \\
CLAM-MB \cite{lu2021data} &
  $93.25_{1.39}$ &
  $96.94_{0.57}$ &
  \multicolumn{1}{c|}{$92.59_{1.51}$} &
  $72.88_{1.56}$ &
  $92.80_{0.32}$ &
  \multicolumn{1}{c|}{$69.43_{1.90}$} &
  $75.22_{2.52}$ &
  $92.83_{1.08}$ &
  $76.09_{2.56}$ \\
DSMIL \cite{li2021dual} &
  $93.48_{1.23}$ &
  $96.67_{0.77}$ &
  \multicolumn{1}{c|}{$92.83_{1.36}$} &
  $69.36_{1.45}$ &
  $92.08_{0.27}$ &
  \multicolumn{1}{c|}{$63.95_{2.69}$} &
  $74.43_{1.88}$ &
  $92.26_{0.85}$ &
  $75.19_{1.80}$ \\
TransMIL \cite{shao2021transmil} &
  $90.51_{5.49}$ &
  $97.02_{0.87}$ &
  \multicolumn{1}{c|}{$89.96_{5.33}$} &
  $67.38_{1.44}$ &
  $91.69_{0.37}$ &
  \multicolumn{1}{c|}{$59.91_{2.61}$} &
  $68.91_{0.73}$ &
  $89.35_{1.09}$ &
  $69.79_{1.14}$ \\
DTFD-MIL \cite{zhang2022dtfd} &
  $93.02_{0.89}$ &
  $97.02_{0.56}$ &
  \multicolumn{1}{c|}{$93.18_{0.91}$} &
  $65.12_{1.23}$ &
  $90.72_{0.34}$ &
  \multicolumn{1}{c|}{$54.40_{3.17}$} &
  $74.14_{2.66}$ &
  $92.68_{1.23}$ &
  $74.32_{2.44}$ \\
ILRA \cite{xiang2023exploring} &
  $91.77_{1.79}$ &
  $95.25_{1.40}$ &
  \multicolumn{1}{c|}{$90.88_{1.99}$} &
  $71.56_{1.25}$ &
  $92.62_{0.16}$ &
  \multicolumn{1}{c|}{$66.70_{1.88}$} &
  $64.17_{4.16}$ &
  $85.85_{1.27}$ &
  $64.44_{5.38}$ \\
AMD-MIL \cite{ling2024agent} &
  $93.33_{1.13}$ &
  $97.48_{0.72}$ &
  \multicolumn{1}{c|}{$92.65_{1.25}$} &
  $73.38_{1.33}$ &
  $93.43_{0.41}$ &
  \multicolumn{1}{c|}{$68.52_{1.74}$} &
  $74.73_{1.55}$ &
  $92.73_{0.81}$ &
  $75.40_{1.30}$ \\
WiKG \cite{li2024dynamic} &
  $93.18_{1.38}$ &
  $97.02_{0.53}$ &
  \multicolumn{1}{c|}{$92.57_{1.42}$} &
  $73.04_{1.03}$ &
  $93.66_{0.36}$ &
  \multicolumn{1}{c|}{$\underline{69.97_{1.67}}$} &
  $74.54_{2.08}$ &
  $91.94_{0.80}$ &
  $75.15_{1.84}$ \\
ACMIL \cite{zhang2024attention} &
  $92.88_{0.76}$ &
  $97.49_{0.54}$ &
  \multicolumn{1}{c|}{$92.28_{0.71}$} &
  $71.48_{1.90}$ &
  $93.00_{0.48}$ &
  \multicolumn{1}{c|}{$65.95_{2.52}$} &
  $74.23_{1.69}$ &		
  $92.60_{1.19}$ &
  $75.35_{1.62}$ \\ \hline
\rowcolor{blue!8}
\textbf{HookMIL-TN} &
  $\underline{93.85_{0.82}}$ &
  $\underline{97.52_{0.45}}$ &
  \multicolumn{1}{c|}{$\underline{93.42_{0.78}}$} &
  $\underline{73.82_{1.15}}$ &
  $\underline{93.78_{0.30}}$ &
  \multicolumn{1}{c|}{$69.35_{1.42}$} &
  $\underline{75.68_{1.48}}$ &
  $\underline{93.05_{0.65}}$ &
  $\underline{76.32_{1.40}}$ \\
\rowcolor{blue!8}
\textbf{HookMIL-Vis} &
  $\mathbf{94.45_{0.68}}$ &
  $\mathbf{97.92_{0.32}}$ &
  \multicolumn{1}{c|}{$\mathbf{93.98_{0.65}}$} &
  $\mathbf{74.85_{0.95}}$ &
  $\mathbf{94.42_{0.28}}$ &
  \multicolumn{1}{c|}{$\mathbf{70.88_{1.25}}$} &
  $\mathbf{76.65_{1.28}}$ &
  $\mathbf{93.68_{0.58}}$ &
  $\mathbf{77.12_{1.20}}$ \\
\hline
\end{tabular}}}
\end{table*}


In this section, we demonstrate the performance of the proposed HookMIL model on four publicly available pathological tissue slide (WSI) datasets and compare it with other state-of-the-art WSI analysis algorithms. Additionally, we conduct extensive ablation experiments to validate the effectiveness of key design elements in our model.

To comprehensively evaluate the proposed HookMIL method, we conducted experiments on four representative pathology datasets: CAMELYON$^{+}$ \cite{ling2025comprehensive}, PANDA \cite{bulten2022artificial}, TISSUENET \cite{lomenie2022can}, and BCNB (ER, PR, HER2)  \cite{xu2021predicting}. Detailed information can be found in Section ~\ref{supp:B} of the Supplementary Material.

\begin{table*}[tbp]
\caption{Performance [\%] of different MIL methods on multiple WSI prediction tasks based on UNI \cite{chen2024towards}.}
\label{table2}
\centering
\resizebox{2\columnwidth}{!}{
\renewcommand{\arraystretch}{1.5}{
\begin{tabular}{lccccccccc}
\hline
\multirow{2}{*}{\textbf{Method}} &
  \multicolumn{3}{c}{\textbf{BCNB-ER \cite{xu2021predicting}}} &
  \multicolumn{3}{c}{\textbf{BCNB-PR \cite{xu2021predicting}}} &
  \multicolumn{3}{c}{\textbf{BCNB-HER2 \cite{xu2021predicting}}} \\ \cline{2-10} 
  &
  \textbf{Accuracy} &
  \textbf{AUC} &
  \textbf{F1-score} &
  \textbf{Accuracy} &
  \textbf{AUC} &
  \textbf{F1-score} &
  \textbf{Accuracy} &
  \textbf{AUC} &
  \textbf{F1-score} \\ \hline
ABMIL \cite{ilse2018attention} &
  $86.67_{1.41}$ &
  $92.17_{1.33}$ &
  \multicolumn{1}{c|}{$78.57_{3.43}$} &
  $82.70_{2.97}$ &
  $84.11_{2.70}$ &
  \multicolumn{1}{c|}{$74.17_{5.91}$} &
  $74.95_{2.23}$ &
  $75.28_{2.20}$ &
  $61.77_{7.60}$ \\
CLAM-SB \cite{lu2021data} &
  $86.19_{2.85}$ &
  $92.41_{1.34}$ &
  \multicolumn{1}{c|}{$76.63_{7.53}$} &
  $82.80_{2.30}$ &
  $85.16_{2.85}$ &
  \multicolumn{1}{c|}{$76.04_{2.95}$} &
  $76.65_{0.73}$ &
  $77.93_{2.19}$ &
  $62.70_{6.65}$ \\
CLAM-MB \cite{lu2021data} &
  $87.14_{1.75}$ &
  $92.48_{1.39}$ &
  \multicolumn{1}{c|}{$79.49_{3.59}$} &
  $82.87_{2.58}$ &
  $84.88_{3.22}$ &
  \multicolumn{1}{c|}{$75.50_{4.78}$} &
  $76.75_{1.64}$ &
  $77.45_{2.08}$ &
  $65.19_{8.99}$ \\
DSMIL \cite{li2021dual} &
  $88.28_{1.89}$ &
  $93.00_{1.53}$ &
  \multicolumn{1}{c|}{$80.38_{4.91}$} &
  $82.51_{2.01}$ &
  $85.18_{2.27}$ &
  \multicolumn{1}{c|}{$74.66_{2.68}$} &
  $77.50_{1.14}$ &
  $76.37_{2.78}$ &
  $65.51_{3.92}$ \\
TransMIL \cite{shao2021transmil} &
  $85.73_{1.83}$ &
  $91.04_{1.57}$ &
  \multicolumn{1}{c|}{$75.41_{5.25}$} &
  $77.59_{5.58}$ &
  $84.23_{2.88}$ &
  \multicolumn{1}{c|}{$71.60_{3.95}$} &
  $74.01_{1.08}$ &
  $75.41_{2.39}$ &
  $52.71_{10.30}$ \\
DTFD-MIL \cite{zhang2022dtfd} &
  $87.43_{2.77}$ &
  $90.78_{1.33}$ &
  \multicolumn{1}{c|}{$80.71_{4.19}$} &
  $81.94_{3.63}$ &
  $84.76_{4.22}$ &
  \multicolumn{1}{c|}{$76.34_{3.75}$} &
  $76.18_{1.61}$ &
  $77.18_{2.49}$ &
  $62.61_{10.12}$ \\
ILRA \cite{xiang2023exploring} &
  $85.53_{1.89}$ &
  $90.24_{1.97}$ &
  \multicolumn{1}{c|}{$75.73_{5.92}$} &
  $81.19_{1.66}$ &
  $83.63_{1.41}$ &
  \multicolumn{1}{c|}{$72.97_{3.00}$} &
  $75.33_{2.15}$ &
  $73.25_{3.14}$ &
  $63.56_{5.57}$ \\
AMD-MIL \cite{ling2024agent} &
  $85.63_{1.42}$ &
  $91.92_{1.86}$ &
  \multicolumn{1}{c|}{$75.22_{6.05}$} &
  $82.70_{1.67}$ &
  $84.93_{3.17}$ &
  \multicolumn{1}{c|}{$75.80_{3.45}$} &
  $77.41_{1.38}$ &
  $77.83_{3.02}$ &
  $63.77_{6.56}$ \\
WiKG \cite{li2024dynamic} &
  $87.99_{1.38}$ &
  $92.61_{1.71}$ &
  \multicolumn{1}{c|}{$\underline{81.99_{2.44}}$} &
  $79.21_{5.60}$ &
  $85.21_{2.62}$ &
  \multicolumn{1}{c|}{$69.67_{10.99}$} &
  $76.65_{1.54}$ &
  $76.75_{2.78}$ &
  $60.31_{9.01}$ \\ 
AC-MIL \cite{zhang2024attention} &
  $86.95_{1.80}$ &
  $91.86_{2.14}$ &
  \multicolumn{1}{c|}{$77.73_{5.57}$} &
  $81.66_{1.83}$ &
  $84.83_{2.84}$ &
  \multicolumn{1}{c|}{$72.29_{2.14}$} &
  $76.75_{1.53}$ &
  $78.80_{2.90}$ &
  $59.94_{10.74}$ \\ \hline
\rowcolor{blue!8}
\textbf{HookMIL-TN} &
  $\underline{88.55_{1.45}}$ &
  $\underline{93.22_{1.25}}$ &
  \multicolumn{1}{c|}{$81.85_{2.85}$} &
  $\underline{83.18_{1.75}}$ &
  $\underline{85.42_{2.20}}$ &
  \multicolumn{1}{c|}{$\underline{76.25_{2.45}}$} &
  $\underline{77.85_{1.15}}$ &
  $\underline{78.95_{2.20}}$ &
  $\underline{65.75_{4.15}}$ \\
\rowcolor{blue!8}
\textbf{HookMIL-Vis} &
  $\mathbf{89.65_{1.18}}$ &
  $\mathbf{94.28_{0.95}}$ &
  \multicolumn{1}{c|}{$\mathbf{83.92_{2.45}}$} &
  $\mathbf{84.87_{1.42}}$ &
  $\mathbf{87.12_{1.86}}$ &
  \multicolumn{1}{c|}{$\mathbf{78.45_{2.15}}$} &
  $\mathbf{79.36_{0.98}}$ &
  $\mathbf{80.95_{1.75}}$ &
  $\mathbf{67.73_{3.62}}$ \\
\hline
\end{tabular}}}
\end{table*}

\begin{figure}[t]
\centering
\includegraphics[width=\linewidth]{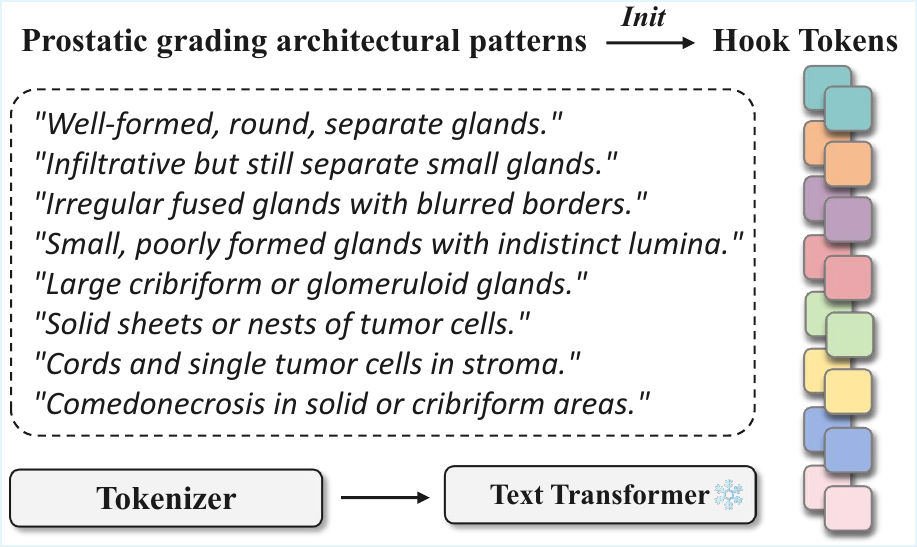}  
\caption{Text-based initialization of hook tokens using prostatic grading architectural patterns.}
\label{fig3}
\end{figure}

\begin{table}[h]
\caption{Classification performance [\%] (AUC) of various methods on PANDA \cite{bulten2022artificial} dataset based on CONCH \cite{lu2024visual}.}
\label{table3}
\centering
\renewcommand{\arraystretch}{1.2}
\small
\begin{tabular}{lccc}
\hline
\textbf{Method} & ACC & AUC & F1 \\
\hline
ABMIL \cite{ilse2018attention}    & $67.39_{0.40}$ & $91.81_{0.27}$ & $61.96_{0.46}$ \\
CLAM-SB \cite{lu2021data}         & $67.61_{0.69}$ & $91.90_{0.26}$ & $62.10_{0.79}$ \\
CLAM-MB \cite{lu2021data}         & $67.95_{0.71}$ & $91.90_{0.17}$ & $62.40_{0.97}$ \\
DSMIL \cite{li2021dual}           & $65.48_{1.06}$ & $90.76_{0.29}$ & $59.56_{1.25}$ \\
TransMIL \cite{shao2021transmil}  & $66.09_{1.04}$ & $91.16_{0.23}$ & $59.81_{0.87}$ \\
DTFD-MIL \cite{zhang2022dtfd}     & $67.62_{0.55}$ & $91.80_{0.18}$ & $61.96_{1.14}$ \\
ILRA \cite{xiang2023exploring}    & $70.02_{1.03}$ & $92.65_{0.27}$ & $65.10_{1.39}$ \\
AMD-MIL \cite{ling2024agent}      & $68.24_{1.61}$ & $92.06_{0.31}$ & $63.12_{1.90}$ \\
WiKG \cite{li2024dynamic}         & $68.58_{0.95}$ & $92.29_{0.35}$ & $63.80_{1.17}$ \\
ACMIL \cite{zhang2024attention}   & $66.58_{0.65}$ & $91.38_{0.17}$ & $60.65_{0.95}$ \\
\hline
\cellcolor{blue!8}\textbf{HookMIL-TN}  & \cellcolor{blue!8}$70.85_{0.82}$ & \cellcolor{blue!8}$92.88_{0.22}$ & \cellcolor{blue!8}$65.45_{1.12}$ \\
\cellcolor{blue!8}\textbf{HookMIL-Vis} & \cellcolor{blue!8}\underline{$71.62_{0.68}$} & \cellcolor{blue!8}\underline{$93.25_{0.18}$} & \cellcolor{blue!8}\underline{$66.28_{0.95}$} \\
\cellcolor{blue!8}\textbf{HookMIL-Txt} & \cellcolor{blue!8}$\mathbf{72.15_{0.59}}$ & \cellcolor{blue!8}$\mathbf{93.52_{0.15}}$ & \cellcolor{blue!8}$\mathbf{67.75_{0.82}}$ \\
\hline
\end{tabular}
\end{table}

\begin{table}[h]
\caption{Biomarker prediction performance [\%] (AUC) of various methods on BCNB \cite{xu2021predicting} datasets based on Loki \cite{chen2025visual}.}
\label{table4}
\centering
\renewcommand{\arraystretch}{1.2}
\small
\begin{tabular}{lccc}
\hline
\textbf{Method} & \textbf{ER \cite{xu2021predicting}} & \textbf{PR \cite{xu2021predicting}} & \textbf{HER2 \cite{xu2021predicting}} \\
\hline
ABMIL \cite{ilse2018attention} & $69.23_{2.32}$ & $67.73_{3.68}$ & $61.36_{5.49}$ \\
CLAM-SB \cite{lu2021data}      & $67.24_{2.14}$ & $66.83_{4.10}$ & $59.57_{4.47}$ \\
CLAM-MB \cite{lu2021data}      & $66.31_{2.42}$ & $66.59_{4.45}$ & $59.73_{4.52}$ \\
DSMIL \cite{li2021dual}        & $62.22_{4.65}$ & $59.15_{2.96}$ & $57.54_{3.98}$ \\
TransMIL \cite{shao2021transmil} & $64.05_{5.01}$ & $62.57_{1.83}$ & $60.07_{2.52}$ \\
DTFD-MIL \cite{zhang2022dtfd}  & $66.94_{2.40}$ & $65.74_{4.20}$ & $58.78_{4.93}$ \\
ILRA \cite{xiang2023exploring} & $60.97_{5.14}$ & $59.72_{3.50}$ & $56.72_{2.91}$ \\
AMD-MIL \cite{ling2024agent}   & $70.42_{3.99}$ & $70.09_{2.33}$ & $63.61_{5.19}$ \\
WiKG \cite{li2024dynamic}      & $69.52_{4.61}$ & $67.45_{3.46}$ & $61.25_{7.37}$ \\
AC-MIL \cite{zhang2024attention} & $69.48_{4.38}$ & $68.60_{2.19}$ & $63.49_{4.88}$ \\
\hline
\cellcolor{blue!8}\textbf{HookMIL-TN}  & \cellcolor{blue!8}\underline{$71.25_{2.85}$} & \cellcolor{blue!8}\underline{$70.88_{1.42}$} & \cellcolor{blue!8}\underline{$64.35_{3.15}$} \\
\cellcolor{blue!8}\textbf{HookMIL-ST}  & \cellcolor{blue!8}$\mathbf{73.18_{2.36}}$ & \cellcolor{blue!8}$\mathbf{72.65_{1.28}}$ & \cellcolor{blue!8}$\mathbf{66.42_{2.68}}$ \\
\hline
\end{tabular}
\end{table}

\subsection{Implementation Details}
During the image preprocessing stage, we applied uniform processing to all four datasets: each WSI was divided into non-overlapping patches of 256$\times$256 pixels at a magnification of 20$\times$. All experiments were conducted in a Python environment using the PyTorch deep learning framework on workstations equipped with NVIDIA RTX A100 GPUs. The image encoders from pre-trained pathology models UNI \cite{chen2024towards}, CONCH \cite{lu2024visual}, and Loki \cite{chen2025visual} were employed as feature encoding models. All experiments employed consistent hyperparameters and settings: training was conducted for 30 epochs using the Adam optimizer with an initial learning rate of $10^{-4}$ and weight decay of $10^{-5}$. Cross-entropy loss was adopted as the loss function.

\subsection{Comparison with State-of-the-Art Methods}
In this study, we report the experimental results of our HookMIL framework on four WSI benchmarks: (CAMELYON$^{+}$)~\cite{ling2025comprehensive}, PANDA~\cite{bulten2022artificial}, TISSUENET~\cite{lomenie2022can}, and BCNB~\cite{xu2021predicting}. We compare against ten representative MIL baselines: ABMIL~\cite{ilse2018attention}, CLAM-SB~\cite{lu2021data}, CLAM-MB~\cite{lu2021data}, DSMIL~\cite{li2021dual}, TransMIL~\cite{shao2021transmil}, DTFD-MIL~\cite{zhang2022dtfd}, ILRA~\cite{xiang2023exploring}, AMD-MIL~\cite{ling2024agent}, WiKG~\cite{li2024dynamic}, and ACMIL~\cite{zhang2024attention}. For fairness, all methods use the same feature inputs and training protocol. As summarized in Table~\ref{table1} and Table~\ref{table2}, HookMIL achieves state-of-the-art performance in terms of accuracy, AUC, and macro F1 on most evaluation settings, highlighting its ability to capture inter-instance context while preserving effective bag-level aggregation. Furthermore, our HookMIL-Vis variant, which initializes hook tokens with key-patch visual representations derived from pathology priors, yields consistently better performance, indicating that prior pathological knowledge enhances the querying capability of hook tokens.

\subsection{Ablation of Hook Diversity Regularization}

To quantify the effect of the proposed hook diversity regularization, we conduct an ablation study by varying the diversity coefficient $\lambda$. As shown in Figure~\ref{fig4}(a), the optimal performance on each dataset–backbone pair is consistently achieved at a non-zero value of $\lambda$. The best results are obtained with $\lambda$ values between 0.1 and 0.7, indicating that a moderate amount of diversity regularization is generally preferred.

Across all dataset–backbone combinations, both $\lambda = 0$ (no diversity constraint) and excessively large $\lambda$ lead to degraded performance, while a moderate level of diversity regularization consistently improves AUC and F1. This trend suggests that the Hook Diversity Loss effectively encourages hooks to encode complementary contextual patterns without overly constraining their representational capacity, thereby enhancing the discriminative power of HookMIL.

\begin{figure*}[h]
\centering
\includegraphics[width=\textwidth]{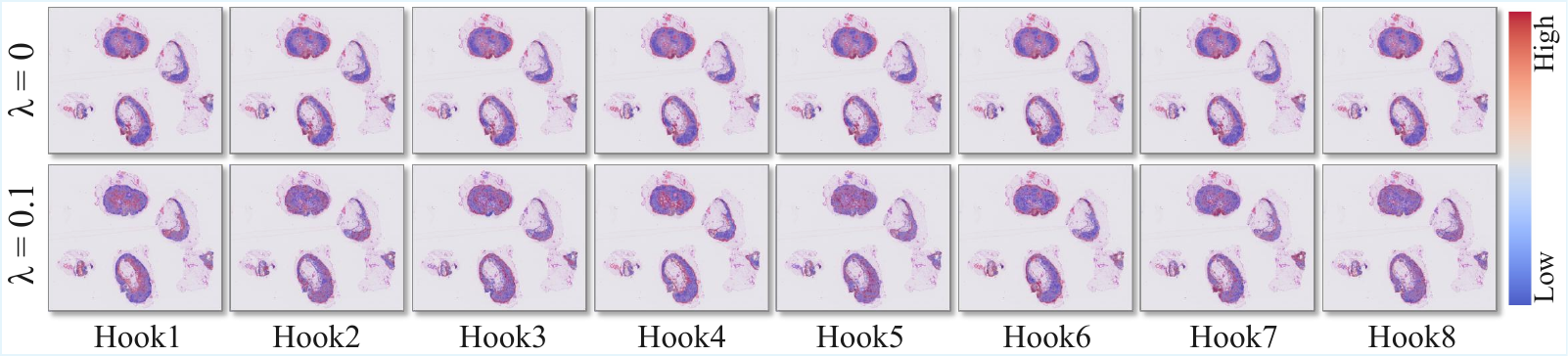}  
\caption{Attention maps of different hook tokens under varying Hook Diversity Loss weights. }
\label{fig5}
\end{figure*}

Beyond quantitative gains, hook diversity also improves interpretability. As shown in Figure~\ref{fig5}, we visualize the attention maps of different hooks under varying $\lambda$ values. When $\lambda = 0$, multiple hooks tend to collapse onto highly similar regions, focusing on the same dominant tumor areas and failing to capture diverse morphological cues. In contrast, with the Hook Diversity Loss activated (e.g., $\lambda = 0.1$), different hooks attend to distinct yet complementary regions, such as tumor cores, invasive margins, scattered micro-foci, and surrounding stromal tissue. This behavior indicates that the regularization encourages hooks to specialize in different contextual patterns, leading to more comprehensive coverage of diagnostically relevant structures. Such specialization not only explains the observed performance gains, but also yields more interpretable attention patterns that align more closely with pathologists’ reading habits.

\subsection{Ablation of Hook Token Numbers}
\begin{figure}[h]
\centering
\includegraphics[width=\linewidth]{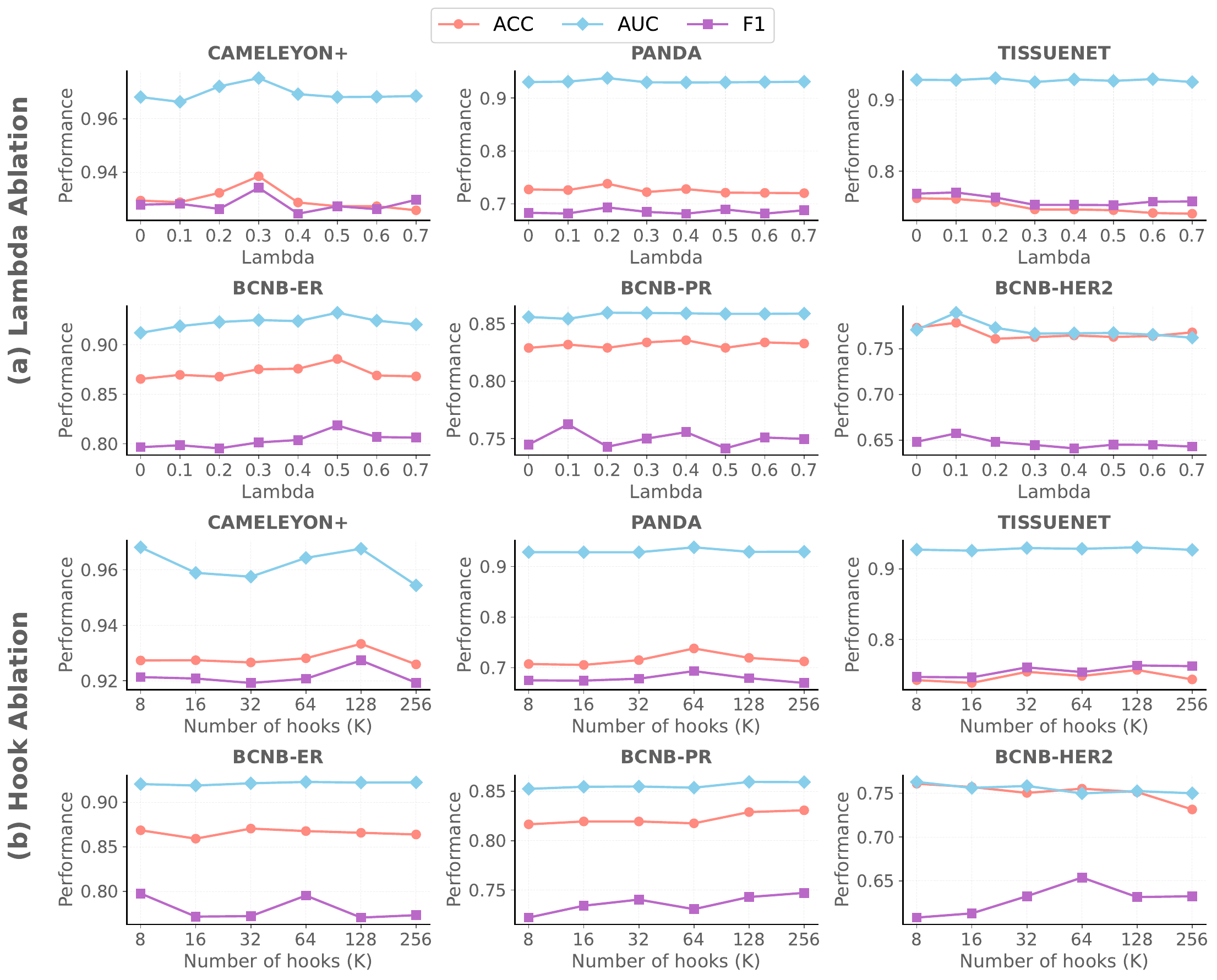}  
\caption{Hyperparameter ablation of HookMIL. (a) Ablation of Hook Diversity Loss. (b) Ablation of Hook Token Numbers.
}
\label{fig4}
\end{figure}

As shown in Figure \ref{fig4} (b), we further ablate the effect of the number of hook tokens while fixing $\lambda = 0.2$. Overall, the model is relatively robust to this hyper-parameter, and increasing the number of hook tokens tends to yield larger performance gains on more complex tasks.

\subsection{Flexible Hook Initialization Methods}
As shown in Table~\ref{table2} and Table~\ref{table3}, to evaluate the flexibility of hook tokens under different pretraining modalities, we conduct two groups of experiments. First, we build on CONCH \cite{lu2024visual}, a pathology-specific vision--language foundation model, and instantiate HookMIL-Txt on the PANDA~\cite{bulten2022artificial} dataset by initializing hook tokens from the aligned text embeddings. In this setting, HookMIL-Txt achieves the best performance among all variants, indicating that language-aligned hooks can effectively inject high-level semantic priors into the MIL aggregation process. Second, we leverage Loki \cite{chen2025visual}, a pathology--genomics foundation model, and construct HookMIL-ST by initializing hook tokens in the latent space shared with gene-level representations. On the corresponding benchmarks, HookMIL-ST attains the strongest results, demonstrating that transcriptomic-aware hooks can further enhance slide-level modeling when suitable molecular priors are available. Nevertheless, due to the limited scale and fragmented nature of current multimodal pathology pretraining data, existing multimodal foundation models still lag behind large-scale vision-only pathology encoders such as UNI~\cite{chen2024towards} in overall representation quality. Our findings, however, suggest that multimodal space alignment offers a promising new route for MIL: as data silos are gradually broken and multimodal pathology foundation models become more mature, hook-based multimodal querying is expected to unlock a higher performance ceiling than purely visual MIL paradigms.

\subsection{Interpretability of HookMIL}
\begin{figure}[t]
\centering
\includegraphics[width=\linewidth]{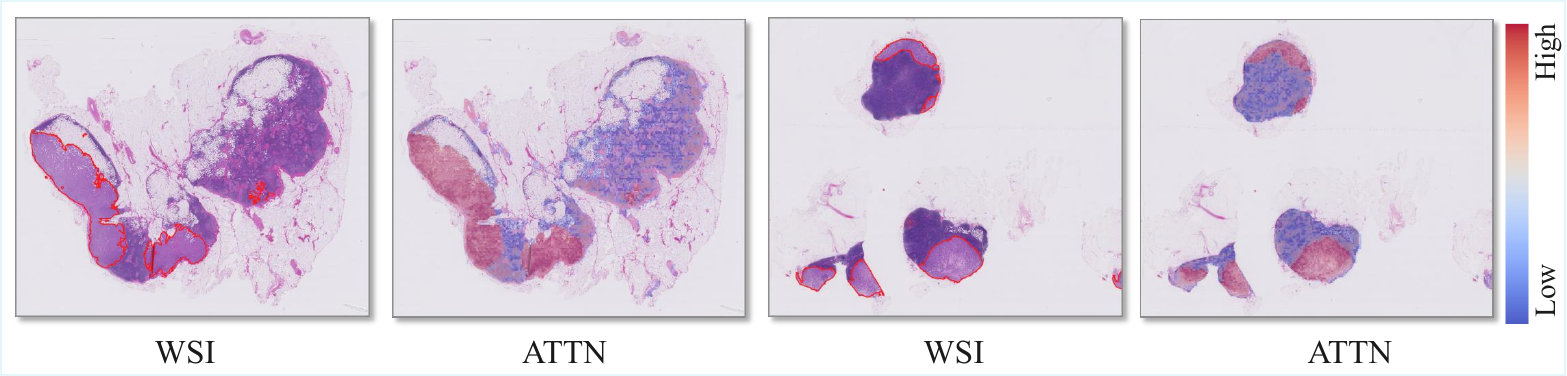}  
\caption{Attention distribution of HookMIL on whole-slide images.}
\label{fig6}
\end{figure}

We assess the interpretability of HookMIL by visualizing attention maps on WSIs and comparing them with cancer-region annotations. As shown in Figure~\ref{fig6}, regions with high hook attention largely coincide with annotated tumor areas, while normal tissue receives low responses. This alignment indicates that HookMIL captures clinically meaningful patterns and provides an intuitive, spatial explanation of its slide-level predictions.

%% file: 5_conclusion.tex
\section{Conclusion and Future Work}
In this paper, we presented HookMIL, a context-aware MIL framework that models inter-instance dependencies via a compact set of learnable hook tokens. With bidirectional hook attention and a diversity regularizer, HookMIL overcomes the limitations of independent instance processing and global self-attention, achieving state-of-the-art performance with improved efficiency and interpretability. Looking ahead, HookMIL suggests several promising directions:

\textbf{Hook-based context modeling}. Replacing self-interactive attention with lightweight hook-style interactions provides a more scalable paradigm for contextual modeling in large-bag MIL.

\textbf{Multimodal integration}. The hook interface naturally aligns with visual and multimodal pathology foundation models, offering a flexible way to inject image, textual, and clinical signals as large-scale multimodal pretraining continues to evolve.

\textbf{Slide-level foundation models}. Compared with TITAN-style \cite{ding2025multimodal} WSI-level models that rely on full self-attention over all instances, hook-based designs can, in principle, support efficient full-slide representation learning by capturing global context without quadratic complexity.

%% file: 6_supp.tex
\clearpage
\renewcommand\thesection{\Alph{section}}
\setcounter{section}{0}
\maketitlesupplementary
\allowdisplaybreaks
\newtheorem{theorem}{Theorem}
\newtheorem{lemma}{Lemma}
\newtheorem{proposition}{Proposition}
\newtheorem{corollary}{Corollary}
\newtheorem{definition}{Definition}
\newtheorem{remark}{Remark}

\section{Theoretical Analysis of HookMIL}

This section provides a compact theoretical perspective on HookMIL from three aspects: (i) notation and one-round update, (ii) a low-rank view of induced inter-instance dependencies, and (iii) gradient propagation properties.

\subsection{Notation and One-Round Update}

Let $X = [x_1,\dots,x_N]^\top \in \mathbb{R}^{N \times D}$ denote instance features (row-wise), and $H = [h_1,\dots,h_K]^\top \in \mathbb{R}^{K \times D}$ denote hook tokens, where $K \ll N$ is the number of hooks.

For clarity, we collect the linear maps used in the two cross-attention directions:
\[
\begin{aligned}
Q_h &= H W_q^\top,     &\quad K_x &= X W_k^\top,     &\quad V_x &= X W_v^\top, \\
Q_x &= X W_q'^\top,    &\quad K_h &= H' W_k'^\top,    &\quad V_h &= H' W_v'^\top,
\end{aligned}
\]
where all $W_{\cdot}$ are learnable matrices in $\mathbb{R}^{D \times D}$ (or absorbed into the tokens/features).

The two attention kernels are
\[
\begin{aligned}
S_{h2x} &= \mathrm{softmax}\!\left(\frac{Q_h K_x^\top}{\sqrt D}\right)
  \in \mathbb{R}^{K \times N}, \\
S_{x2h} &= \mathrm{softmax}\!\left(\frac{Q_x K_h^\top}{\sqrt D}\right)
  \in \mathbb{R}^{N \times K},
\end{aligned}
\]
with row-wise softmax.

A single HookMIL round can then be summarized as:
\[
\begin{aligned}
\tilde H &= \mathrm{LN}\!\left(H + S_{h2x} V_x\right), \\
H' &= \mathrm{LN}\!\left(\tilde H + \mathrm{MHSA}(\tilde H)\right), \\
X' &= \mathrm{LN}\!\left(X + S_{x2h} V_h\right),
\end{aligned}
\]
where $\mathrm{MHSA}$ denotes multi-head self-attention over the $K$ hooks, and $\mathrm{LN}$ is LayerNorm. The term $S_{x2h} V_h$ can be viewed as a context operator that injects hook-mediated information back into instances.

\subsection{Low-Rank View of Inter-Instance Dependencies}
\label{supp_low_rank}

We now show that HookMIL induces a \emph{low-rank} instance–instance dependency structure via hooks.

\begin{proposition}[Rank-$K$ attention operator]
\label{prop:rankK}
Consider the HookMIL context operator
\[
\mathcal{T}_{\mathrm{Hook}}(X) := S_{x2h} V_h \in \mathbb{R}^{N \times D}.
\]
Without loss of generality, the value transformation $W_v'$ can be absorbed into the hook representation so that
\[
\mathcal{T}_{\mathrm{Hook}}(X) = S_{x2h} U_h,
\quad U_h \in \mathbb{R}^{K \times D}.
\]
The induced instance-to-instance dependency matrix
\[
\widetilde A := S_{x2h} S_{h2x} \in \mathbb{R}^{N \times N}
\]
satisfies
\[
\mathrm{rank}(\widetilde A) \le K.
\]
\end{proposition}

\begin{proof}
Since $S_{x2h} \in \mathbb{R}^{N \times K}$ and $S_{h2x} \in \mathbb{R}^{K \times N}$, 
the product $\widetilde{A} = S_{x2h} S_{h2x}$ is a multiplication between 
an $N \times K$ matrix and a $K \times N$ matrix, thus its rank is at most $K$.

\end{proof}

\noindent
In contrast, full self-attention on $N$ instances constructs a dense
$N \times N$ kernel with rank up to $N$. Proposition~\ref{prop:rankK}
shows that HookMIL can be interpreted as learning a rank-$K$ factorization
of the global dependency structure through a set of hook ``bases''.
Each hook corresponds to a basis vector in the induced subspace, and
the matrices $S_{h2x}$ and $S_{x2h}$ describe how instances are
projected onto and reconstructed from these bases.

\begin{remark}[Diversity and basis incoherence]
Although the diversity regularization is defined in the main text,
its effect can be intuitively understood here: encouraging different
hooks to attend to complementary regions reduces the coherence between
rows of $S_{h2x}$, which in turn improves the conditioning and expressive
coverage of the low-rank operator $\widetilde A = S_{x2h} S_{h2x}$.
\end{remark}

\subsection{Gradient Propagation Properties}
\label{supp_grid}

We next analyze how gradients propagate through the hook-mediated
architecture, highlighting that HookMIL preserves efficient
long-range communication without explicit $N^2$ coupling.

Let $\mathcal{L}$ be the training loss and
$G = \partial \mathcal{L}/\partial X' \in \mathbb{R}^{N \times D}$.
From the update
\[
X' = X + S_{x2h} V_h,
\]
we obtain the gradient w.r.t.\ $X$.

\begin{lemma}[Gradient to instances]
\label{lem:gradX}
The gradient to $X$ can be decomposed as
\[
\frac{\partial \mathcal{L}}{\partial X}
= G + \frac{\partial(S_{x2h} V_h)}{\partial X} : G,
\]
where ``$:$'' indicates tensor contraction under the chain rule.
Using the Jacobian of row-wise softmax,
$J_{\mathrm{softmax}}(u) = \mathrm{Diag}(p) - p p^\top$ for $p = {\operatorname{softmax}}(u)$,
one can bound
\[
\begin{aligned}
\bigg\|
\frac{\partial \mathcal{L}}{\partial X} - G
\bigg\|_2
&\le
\|J_{S_{x2h}}\|_2
\cdot \|G V_h^\top\|_2
\cdot \|W_q'\|_2
\cdot \|K_h\|_2, \\
\|J_{S_{x2h}}\|_2 &\le \tfrac{1}{4},
\end{aligned}
\]
where $\|\cdot\|_2$ denotes the spectral norm.
\end{lemma}

\begin{proof}
Apply the chain rule to
$S_{x2h} = \mathrm{softmax}(Q_x K_h^\top / \sqrt D)$, then compose with
the linear maps in $Q_x$ and $V_h$. The spectral bound on
$J_{\operatorname{softmax}}$ follows from standard softmax Lipschitz properties.
\end{proof}

\noindent
Similarly, we can characterize how gradients reach the hooks.

\begin{lemma}[Gradient to hooks]
\label{lem:gradH}
Ignoring the 1-Lipschitz constants of LayerNorm and residual connections,
the gradient w.r.t.\ $H$ can be schematically expressed as
\[
\begin{aligned}
\frac{\partial \mathcal{L}}{\partial H}
\approx\
& S_{x2h}^\top G\, W_v'^\top
   \cdot \frac{\partial H'}{\partial H} \\
&+ \left[J_{S_{h2x}} : (\text{terms at } \tilde H)\right]
   \cdot \frac{\partial(Q_h K_x^\top)}{\partial H},
\end{aligned}
\]
where the first term corresponds to the feedback path
$X' \to H' \to H$, and the second term captures how changes in $H$
alter the hook-to-instance aggregation via $S_{h2x}$.
\end{lemma}

\begin{proof}
Again, apply the chain rule through
$X'$, $H'$, and $\tilde H$, noting that $H$ appears both in the
forward feedback path and in the hook-to-instance attention.
\end{proof}

\begin{theorem}[Bidirectional connectivity with path length $\le 2$]
\label{thm:path}
For any pair of instances $(i,j)$, if there exists a hook $k$ such that
$(S_{h2x})_{k,i} (S_{x2h})_{j,k} \neq 0$, then gradients can flow from
instance $i$ to instance $j$ through the path
\[
x_i \;\longrightarrow\; h_k \;\longrightarrow\; x_j
\]
in at most two steps. Consequently, the induced matrix
$\widetilde A = S_{x2h} S_{h2x}$ enables effective global
information propagation across instances while avoiding explicit
$N^2$-size attention.
\end{theorem}

\begin{proof}
Non-zero entries $(S_{h2x})_{k,i}$ and $(S_{x2h})_{j,k}$ imply that
instance $i$ contributes to hook $k$ in the aggregation phase, and
hook $k$ in turn contributes to instance $j$ in the feedback phase.
This establishes a gradient path $i \to k \to j$ with length at most two. Summing over
all hooks provides global connectivity mediated by the rank-$K$ operator
$\widetilde A$.
\end{proof}

Overall, these results show that HookMIL implements a structured
low-rank approximation of global self-attention, where a small number
of hooks both summarizes and redistributes information, while still
maintaining efficient long-range gradient propagation across all
instances.

\section{Dataset Details}
\label{supp:B}

We evaluate HookMIL on four representative whole slide image (WSI) benchmarks spanning different organs and diagnostic settings.

\textbf{CAMELYON$^{+}$} is a breast cancer lymph node metastasis dataset obtained by combining CAMELYON16 and CAMELYON17. It contains 1,349 WSIs in total, including 870 negative slides and 479 positive slides.

\textbf{PANDA} is a large-scale dataset for prostate cancer Gleason grading, comprising 5,455 WSIs. Each slide is labeled with an ISUP grade derived from the Gleason score, resulting in six categories: 1,924 G0, 1,814 G1, 668 G2, 317 G3, 481 G4, and 251 G5.

\textbf{TISSUENET} consists of 1,013 WSIs from cervical biopsies and conizations. The dataset covers a spectrum of cervical lesions with four classes: 268 normal or subnormal cases, 288 LSIL (low-grade squamous intraepithelial lesion), 238 HSIL (high-grade squamous intraepithelial lesion), and 219 invasive squamous carcinomas.

\textbf{BCNB} is an early-stage breast cancer core-needle biopsy dataset containing 1,058 WSIs. From this cohort, we derive three biomarker prediction tasks: ER (BCNB-ER), PR (BCNB-PR), and HER2 (BCNB-HER2), where each slide is annotated with the corresponding immunohistochemical status.

\input{HookMIL}

%% file: HookMIL.tex
\definecolor{myteal}{RGB}{122,165,165}
\algrenewcommand\algorithmiccomment[1]{\hfill\textcolor{gray}{// #1}}

\begin{algorithm*}[ht]
\caption{HookMIL: Revisiting Context Modeling in Multiple Instance Learning for
Computational Pathology}
\label{alg:algorithm1}
\small
\begin{algorithmic}[1]
\Require Training set of WSIs $\{X^l\}_{l=1}^L$ with labels $\{Y_l\}_{l=1}^L$ ($Y_l \in \{0,1,2,\ldots,C-1\}$), number of hook tokens $K$, feature dimension $D$, learning rate $\eta$, diversity coefficient $\lambda$, maximum epochs $E_{max}$
\Ensure Optimal model parameters $\Theta^*$ that minimize $\mathcal{L}_{\text{total}} = \mathcal{L}_{\text{CE}} + \lambda \mathcal{L}_{\text{div}}$

\State Initialize model parameters $\Theta$ and hook tokens $H \in \mathbb{R}^{K \times D} \sim \mathcal{TN}(0, 0.02^2)$
\For{epoch $e = 1$ to $E_{max}$}
    \For{each WSI $X^l$ in training set}
        \State Extract patch features $X \in \mathbb{R}^{N \times D}$ via pretrained encoder
        \State // \textcolor{myteal}{Stage 1: Hook-to-Instance Aggregation}
        \State // Hooks (H) query the instances (X) to gather relevant patterns
        \State $Q_h = H W_q^\top$, $K_x = X W_k^\top$, $V_x = X W_v^\top$
        \State $\mathbf{L} = \frac{Q_h K_x^\top}{\sqrt{D}} \in \mathbb{R}^{K \times N}$ \Comment{Save pre-softmax logits for diversity}
        \State $\mathbf{A}_{h2x} = \text{softmax}(\mathbf{L})$ \Comment{Hook-to-instance attention}
        \State $\tilde{H} = \text{LayerNorm}(H + \mathbf{A}_{h2x} V_x)$ \Comment{Capture local morphological context}

        \State // \textcolor{myteal}{Stage 2: Hook Intercommunication}
        \State // Allow the K hooks to exchange and refine their captured information
        \State $\mathbf{H}_{\text{inter}} = \text{MHSA}(\tilde{H}, \tilde{H}, \tilde{H})$
        \Comment{Exchange global contextual information}
        \State $H' =  \text{LayerNorm}(\tilde{H} +\mathbf{H}_{\text{inter}})$ \Comment{Create refined, context-aware hook tokens}

        \State // \textcolor{myteal}{Stage 3: Instance-to-Hook Feedback}
        \State // Instances (X) query the refined hooks (H') to get contextual updates
        \State $Q_x = X W_q'^\top$, $K_h = H' W_k'^\top$, $V_h = H' W_v'^\top$
        \Comment{Refined Hooks as Keys and Values}
        \State $\mathbf{A}_{x2h} = \text{softmax}(\frac{Q_x K_h^\top}{\sqrt{D}})$
        \State $X' = \text{LayerNorm}(X + \mathbf{A}_{x2h} V_h)$ \Comment{Inject context back into instances}

        \State // \textcolor{myteal}{Stage 4: Diversity Regularization}
        \State // Use the raw logits L from Stage 1 to ensure hooks focus on different patterns
        \State $\mathbf{L}_{\text{norm}} = \frac{\mathbf{L}}{\|\mathbf{L}\|_F + \epsilon}$ \Comment{L2-normalize logits}
        \State $\mathbf{S} = \mathbf{L}_{\text{norm}}\mathbf{L}_{\text{norm}}^\top$ \Comment{Compute KxK similarity matrix of hook patterns}
        \State $\mathcal{L}_{\text{div}} = \frac{1}{K(K-1)}\sum_{i=1}^K \sum_{j \neq i} S_{ij}^2$ \Comment{Penalize redundant hooks}

        \State \textcolor{myteal}{// Stage 5: Bag-level Aggregation and Prediction}
        \State // Apply attention-based pooling on the enhanced instance features $X'$
        \State $a_i = \frac{\exp(w_a^\top \tanh(V_a x'_i))}{\sum_{j=1}^N \exp(w_a^\top \tanh(V_a x'_j))}$ \Comment{Compute attention weights}
        \State $z = \sum_{i=1}^N a_i x'_i$ \Comment{Bag-level representation}
        \State $\hat{y} = \text{softmax}(W_c z + b_c)$
        \State $\mathcal{L}_{\text{CE}} = \mathcal{L}_{\text{ce}}(\hat{y}, Y_l) $
        
        \State // --- \textcolor{myteal}{Loss Computation and Parameter Update} ---
        \State $\mathcal{L}_{\text{total}} = \mathcal{L}_{\text{CE}} + \lambda \mathcal{L}_{\text{div}}$
        \State Update parameters $\Theta$ using gradient descent on $\mathcal{L}_{\text{total}}$ via Adam optimizer
    \EndFor
\EndFor
\State \Return Optimal model parameters $\Theta^*$
\end{algorithmic}
\end{algorithm*}